\documentclass[11pt]{article}

\usepackage[a4paper,margin=1in]{geometry}
\usepackage{amsmath,amssymb,amsfonts}
\usepackage{booktabs}
\usepackage{graphicx}
\usepackage{multirow}
\usepackage{microtype}
\usepackage{xcolor}
\usepackage{url}
\usepackage[hidelinks]{hyperref}
\usepackage[numbers,sort&compress]{natbib}

\graphicspath{{figures/}}

\title{SKIA: Spatial-Knowledge Informed Attention for Robust Remote Photoplethysmography under Lower-Face Motion Artifacts}

\author{Van-Truong Le$^{a,b}$ \and Viet-Tham Huynh$^{a,b}$ \and Minh-Triet Tran$^{a,b}$\\
\small $^a$University of Science, VNU-HCM, Ho Chi Minh City, Vietnam\\
\small $^b$Vietnam National University, Ho Chi Minh City, Vietnam}

\date{}

\begin{document}
\maketitle

\begin{abstract}
Remote photoplethysmography (rPPG) enables contactless estimation of physiological signals from facial videos and has attracted increasing interest in affective computing, healthcare, and human-centered artificial intelligence. However, facial motion remains one of the primary challenges limiting robust physiological reconstruction, particularly during cognitive and communicative tasks where spontaneous mouth and jaw movements introduce severe temporal artifacts. Existing approaches typically model global facial dynamics or employ complex motion-compensation strategies, yet the contribution of lower-face motion to signal degradation remains insufficiently investigated. In this work, we present SKIA, a lightweight, mask-guided framework for motion-robust rPPG estimation, based on the observation that lower-face regions disproportionately contribute to task-induced motion artifacts while providing comparatively limited physiological information. We first perform a systematic analysis of motion-sensitive facial regions using the UBFC-rPPG benchmark and demonstrate that suppressing the lower facial area significantly improves physiological signal reconstruction. Based on this insight, we introduce a static spatial-knowledge informed masking strategy that attenuates motion-dominant regions while preserving stable hemodynamic cues from the forehead and cheeks. Building upon this spatial prior, we further propose SKIA-Net, a compact spatiotemporal architecture that jointly exploits raw appearance information, masked facial representations, and temporal motion residuals for robust blood volume pulse (BVP) reconstruction. Extensive experiments show that the proposed method consistently improves waveform fidelity and heart-rate estimation accuracy over conventional CHROM baselines and single-stream deep models. Static masking improves CHROM correlation from 0.3507 to 0.4166 with a statistically significant gain ($p<0.001$), while the full SKIA-Net achieves a Pearson correlation of 0.7522 and a heart-rate mean absolute error of 3.73 bpm on UBFC-rPPG. Additional ablation studies demonstrate the effectiveness of mask-guided representations and reveal that static anatomical priors outperform dynamic optical-flow-based masking strategies for task-induced motion suppression. Our code is available at \url{https://github.com/lee-vtruong/SKIA-rPPG}.
\end{abstract}


\section{Introduction}
Remote photoplethysmography (rPPG) aims to estimate physiological signals such as blood volume pulse (BVP) and heart rate directly from facial videos without requiring physical contact sensors. By analyzing subtle skin color variations caused by cardiovascular activity, rPPG has emerged as a promising technology for contactless physiological monitoring in applications including telemedicine, affective computing, driver monitoring, human-computer interaction, and behavioral analysis~\citep{verkruysse2008remote}. Compared with conventional wearable sensors, camera-based physiological sensing offers substantial advantages in scalability, convenience, and unobtrusive deployment.

Despite recent progress in classical signal-processing methods and deep learning approaches, robust physiological reconstruction under realistic facial motion remains a major challenge for rPPG systems. Variations caused by head motion, illumination fluctuation, facial expression changes, and speaking activity often dominate the weak hemodynamic signal captured by consumer-grade cameras. Among these factors, lower-face motion induced by active tasks such as speaking, mathematical problem-solving, or cognitive stress is particularly problematic. Continuous mouth articulation, jaw movement, and local non-rigid facial dynamics generate strong temporal artifacts that significantly corrupt pulse estimation.

Existing methods typically address motion artifacts through temporal filtering, global facial stabilization, attention mechanisms, or optical-flow-based compensation. Classical approaches such as CHROM~\citep{dehaan2013robust} and POS~\citep{wang2017algorithmic} rely on handcrafted chrominance projections to suppress illumination variations but remain sensitive to local facial motion. More recent deep learning frameworks leverage convolutional or spatiotemporal architectures to learn physiological representations from video sequences. Although these models achieve promising performance on benchmark datasets, they often treat the facial region uniformly and do not explicitly investigate which facial areas contribute most strongly to motion-induced corruption.

In this work, we revisit the role of spatial facial regions in rPPG estimation and propose a simple observation: during speaking-related activities, the lower facial region introduces disproportionately large motion artifacts while contributing relatively limited stable physiological information compared with the forehead and upper cheek regions. We systematically analyze the effect of suppressing lower-face motion on physiological signal reconstruction.

We first conduct experiments using the UBFC-rPPG benchmark to evaluate conventional CHROM-based rPPG extraction under different masking configurations. We observe that a simple, static lower-face suppression strategy consistently improves physiological waveform correlation and heart-rate estimation accuracy across most subjects. In contrast, dynamic optical-flow-based masking strategies substantially degrade performance due to unstable temporal suppression and excessive signal attenuation. These findings suggest that stable anatomical priors may be more reliable than dynamically estimated motion masks for practical motion-robust rPPG reconstruction.

Based on these observations, we introduce SKIA (Static Kinematic-Informed Attention), a lightweight framework designed to improve physiological robustness in the presence of lower-face motion artifacts. The proposed approach combines static spatial masking with a compact spatiotemporal neural architecture, termed SKIA-Net, that jointly leverages raw facial appearance, masked representations, and temporal motion residuals. Rather than discarding motion information entirely, the model learns complementary representations from both original and motion-suppressed inputs, enabling more stable BVP reconstruction under challenging motion conditions.

Our contributions are summarized as follows:
\begin{itemize}
    \item We provide a systematic analysis of lower-face motion artifacts in rPPG and demonstrate that suppressing motion-dominant facial regions improves physiological reconstruction quality.
    \item We propose a lightweight static kinematic-informed masking strategy that improves CHROM-based rPPG estimation with statistically significant gains across the UBFC-rPPG benchmark.
    \item We introduce SKIA-Net, a compact spatiotemporal architecture that jointly exploits raw appearance, masked representations, and temporal motion cues for robust BVP reconstruction.
    \item We conduct experiments and ablation studies demonstrating the effectiveness of the proposed framework in waveform reconstruction, heart-rate estimation, runtime efficiency, and motion robustness.
\end{itemize}

\section{Related Work}
\subsection{Classical Remote Photoplethysmography Methods}
Early rPPG methods relied on handcrafted signal processing techniques to recover physiological signals from facial videos. Initial approaches commonly estimated pulse signals through spatial color averaging, particularly using the green channel due to its physiological sensitivity to hemoglobin absorption. Subsequent studies introduced blind source separation methods such as Independent Component Analysis and Principal Component Analysis~\citep{poh2010noncontact}. To improve robustness, chrominance-based methods were proposed. CHROM projects normalized RGB signals into a chrominance subspace designed to suppress motion-induced color distortion~\citep{dehaan2013robust}. Later, POS defined a projection plane orthogonal to the skin-tone direction in normalized RGB space~\citep{wang2017algorithmic}.

These classical methods remain computationally efficient and widely used as strong rPPG baselines. However, they rely heavily on spatial color aggregation over selected facial regions. Consequently, region-of-interest selection and motion-contaminated facial areas directly influence reconstructed physiological signals. Our work investigates whether suppressing lower-face motion can improve the stability of chrominance-based reconstruction.

\subsection{Deep Learning for rPPG Estimation}
DeepPhys introduced an early end-to-end convolutional attention framework for camera-based physiological measurement~\citep{chen2018deepphys}. PhysNet adopted a 3D convolutional encoder-decoder architecture for direct BVP waveform reconstruction~\citep{yu2019remote}. Subsequent studies explored efficient architectures such as EfficientPhys~\citep{liu2023efficientphys}, transformer-based temporal modeling~\citep{yu2022physformer}, frequency-domain supervision~\citep{spetlik2018visual}, and self-supervised learning~\citep{gideon2021estimating}.

Although deep learning approaches often outperform traditional pipelines, motion robustness remains a challenge. Most existing methods attempt to learn motion invariance implicitly from training data without explicitly modeling anatomically unstable facial regions. In contrast, SKIA-Net incorporates an explicit lower-face suppression prior into the input representation and combines spatially masked facial representations with temporal physiological modeling.

\subsection{Motion Robustness and Region-of-Interest Selection}
Motion artifacts are among the primary challenges in practical rPPG estimation. Illumination fluctuations, head movement, facial expressions, and camera instability may introduce temporal variations substantially larger than the underlying physiological signal. Previous studies have explored facial stabilization, skin segmentation, adaptive ROI tracking, optical-flow compensation, temporal filtering, and spatial attention. Physiologically informative regions such as the forehead and cheeks are generally more stable than highly deformable facial areas~\citep{lempe2013roi}.

Recent surveys emphasize that real-world robustness remains an open challenge~\citep{chen2024deep}. Most robustness-oriented methods focus on global motion compensation or illumination normalization. In contrast, our work investigates a simpler region-specific prior based on suppressing lower-face motion artifacts. The experiments show that static lower-face suppression improves physiological reconstruction, while dynamic optical-flow-based masking often introduces instability and performance degradation.

\subsection{Datasets and Evaluation Protocols}
UBFC-rPPG is one of the most widely adopted benchmarks for camera-based physiological reconstruction~\citep{bobbia2019unsupervised}. It contains facial videos together with synchronized ground-truth physiological signals acquired using contact sensors. The dataset includes realistic facial motion and physiological variability induced by controlled tasks such as mathematical games and time-sensitive activities. Following prior literature, we evaluate waveform reconstruction using Pearson correlation and heart-rate estimation using mean absolute error (MAE). We further conduct statistical significance analysis, runtime benchmarking, and ablation studies.

\section{Method}
\subsection{Problem Formulation}
Given an input facial video clip $X\in\mathbb{R}^{T\times H\times W\times 3}$, where $T$, $H$, and $W$ denote the number of frames, height, and width, respectively, the goal is to reconstruct the underlying BVP signal $y\in\mathbb{R}^{T}$. Our model predicts a physiological waveform $\hat{y}\in\mathbb{R}^{T}$. Reconstruction quality is assessed using waveform-level Pearson correlation and heart-rate estimation accuracy~\citep{liu2023rppgtoolbox}.

The Pearson correlation coefficient is
\begin{equation}
\rho(\hat{y},y)=\frac{\sum_t(\hat{y}_t-\bar{\hat{y}})(y_t-\bar{y})}{\sqrt{\sum_t(\hat{y}_t-\bar{\hat{y}})^2}\sqrt{\sum_t(y_t-\bar{y})^2}},
\end{equation}
where $\bar{\hat{y}}$ and $\bar{y}$ denote temporal means. For heart-rate estimation, the dominant frequency peak is extracted from the reconstructed BVP spectrum using the Fast Fourier Transform (FFT) and converted to beats per minute. Performance is measured by
\begin{equation}
\mathrm{MAE}=\frac{1}{N}\sum_{i=1}^{N}\left|\mathrm{HR}^{i}_{\mathrm{pred}}-\mathrm{HR}^{i}_{\mathrm{gt}}\right|.
\end{equation}

\subsection{Static Lower-Face Suppression and Masked rPPG}
\begin{figure}[t]
    \centering
    \includegraphics[width=0.82\linewidth]{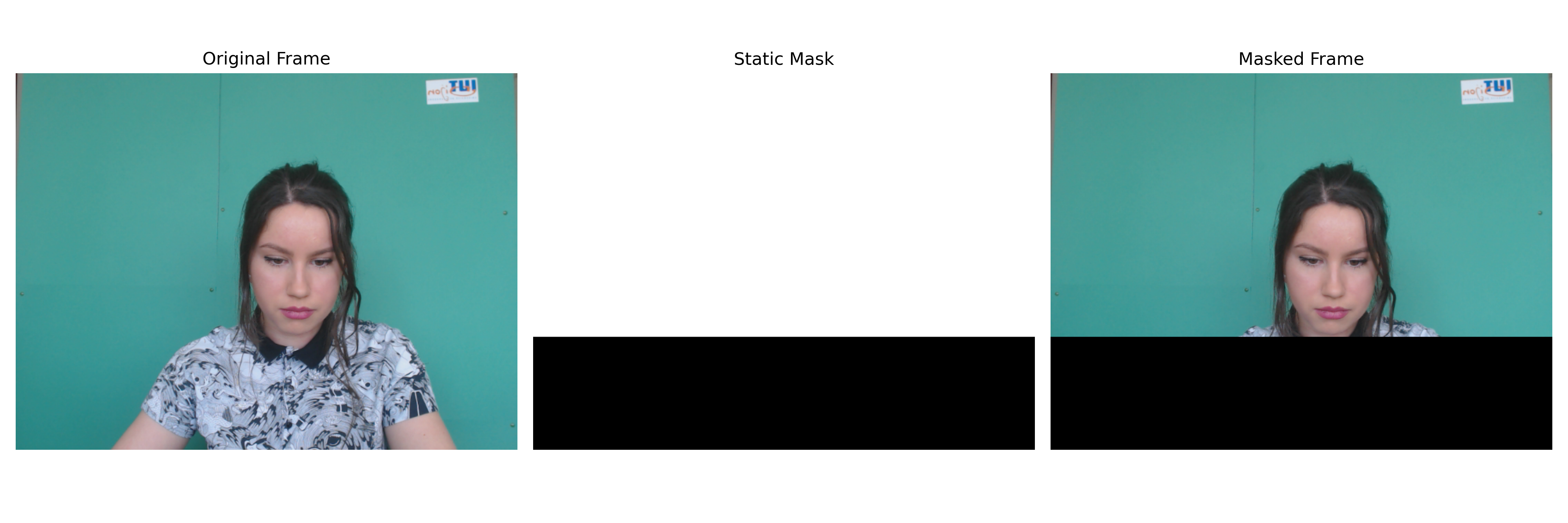}
    \caption{Example of static lower-face suppression. The lower facial region is removed while preserving relatively stable physiological regions such as the forehead and cheeks.}
    \label{fig:mask}
\end{figure}

Lower-face regions introduce strong motion artifacts during speaking-related activities because mouth articulation, jaw deformation, and non-rigid motion generate large temporal intensity fluctuations. We define a static spatial mask $M\in\mathbb{R}^{H\times W}$ that preserves the upper facial area while attenuating the lower portion:
\begin{equation}
M(i,j)=
\begin{cases}
1, & i<\alpha H,\\
0, & \text{otherwise},
\end{cases}
\end{equation}
where $\alpha$ controls the retained facial ratio. We empirically set $\alpha=0.7$. The masked representation is $X_{\mathrm{masked}}=X\odot M$, where $\odot$ denotes element-wise multiplication. Figure~\ref{fig:mask} illustrates the strategy.

We evaluate the influence of $\alpha$ in an ablation study. Smaller values remove excessive physiologically informative skin regions, whereas larger values fail to sufficiently suppress motion-contaminated mouth regions. Unlike dynamic motion compensation, the proposed mask remains temporally stable across all frames.

\paragraph{Masked CHROM.} Conventional CHROM extracts temporal RGB traces by averaging color values over the facial region. For a frame $X_t$, standard aggregation is $R_t=\frac{1}{HW}\sum_{i,j}X^R_{t,i,j}$, and analogously for $G_t$ and $B_t$. Masked CHROM instead computes weighted spatial averages:
\begin{equation}
R_t=\frac{\sum_{i,j}M_{i,j}X^R_{t,i,j}}{\sum_{i,j}M_{i,j}},
\end{equation}
with analogous expressions for $G_t$ and $B_t$. The normalized chrominance projections are
\begin{equation}
X_{\mathrm{chr}}=0.77R-0.51G,\qquad Y_{\mathrm{chr}}=0.77R+0.51G-0.77B,
\end{equation}
followed by bandpass filtering in the heart-rate frequency range.

\paragraph{Dynamic mask attempt.} We also explored optical-flow-based dynamic masking, where dense optical flow between consecutive frames generates motion-dependent masks~\citep{farneback2003two}. These dynamic masks substantially degraded CHROM and deep reconstruction performance because frame-wise mask fluctuations introduced artificial temporal color variations and interfered with stable pulse extraction.

\subsection{SKIA-Net Architecture and Training}
\begin{figure}[t]
    \centering
    \includegraphics[width=\linewidth]{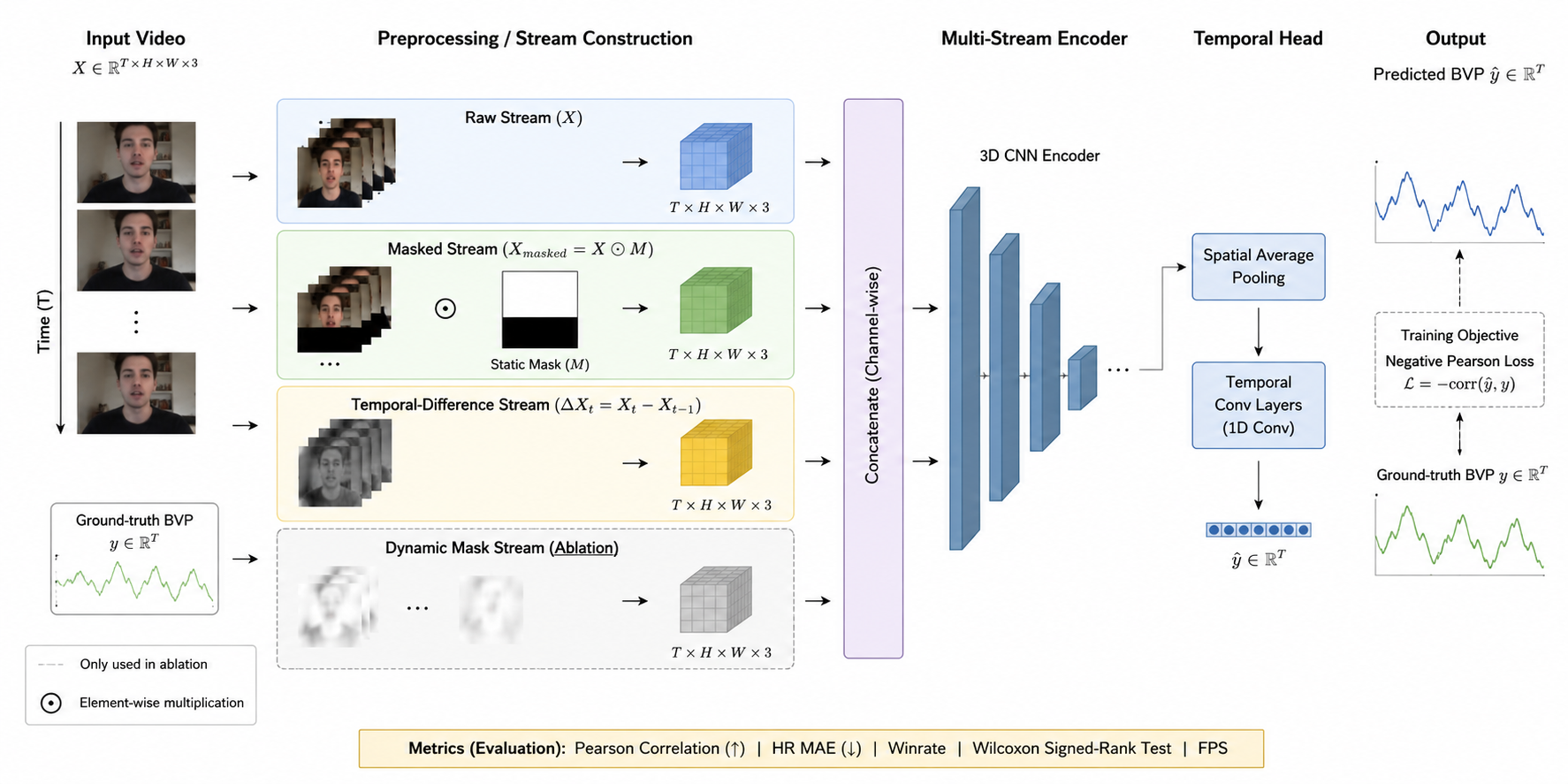}
    \caption{Overview of SKIA. Raw frames, static lower-face suppressed frames, and temporal-difference frames are jointly encoded to reconstruct the target BVP waveform.}
    \label{fig:pipeline}
\end{figure}

SKIA-Net jointly exploits three complementary representations: the raw appearance stream $X$, the static masked stream $X_{\mathrm{masked}}=X\odot M$, and the temporal-difference stream $\Delta X_t=X_t-X_{t-1}$. These representations are concatenated along the channel dimension and processed by a lightweight 3D convolutional encoder. Spatial average pooling followed by a temporal convolutional prediction head reconstructs the target BVP waveform. Figure~\ref{fig:pipeline} presents the architecture.

Table~\ref{tab:variants} summarizes the evaluated input configurations.

\begin{table}[t]
\centering
\caption{SKIA-Net variants and input configurations.}
\label{tab:variants}
\begin{tabular}{lccc}
\toprule
Variant & Raw & Masked & Temporal diff.\\
\midrule
SKIA Raw & $\checkmark$ & & \\
SKIA Masked & & $\checkmark$ & \\
SKIA Raw+Masked & $\checkmark$ & $\checkmark$ & \\
SKIA Full & $\checkmark$ & $\checkmark$ & $\checkmark$\\
\bottomrule
\end{tabular}
\end{table}

The network is trained using Negative Pearson Correlation Loss:
\begin{equation}
\mathcal{L}=-\frac{\sum_t(\hat{y}_t-\bar{\hat{y}})(y_t-\bar{y})}{\sqrt{\sum_t(\hat{y}_t-\bar{\hat{y}})^2}\sqrt{\sum_t(y_t-\bar{y})^2}}.
\end{equation}
Compared with point-wise regression losses such as MSE, Pearson-based optimization focuses on waveform consistency and temporal alignment.

\section{Experiments and Results}
\subsection{Dataset}
All experiments use UBFC-rPPG. The dataset contains facial videos from 42 subjects performing controlled tasks involving natural facial motion and physiological variation. Each sample provides an RGB facial video sequence, a ground-truth BVP waveform, and instantaneous heart-rate measurements from a contact sensor.

We split the dataset subject-wise into 32 training subjects and 10 testing subjects. The fixed test set is listed as subject43, subject44, subject45, subject46, subject47, subject48, subject49, subject5, subject8, and subject9; the remaining subjects form the training set. Video clips are segmented into fixed-length windows of 64 or 128 frames, with stride equal to the clip length or half-overlap. Frames are resized to 96$\times$96 or 128$\times$128 pixels. Ground-truth and predicted signals are bandpass filtered within 0.7--2.5 Hz (approximately 42--150 bpm).

\subsection{Implementation Details}
Experiments were implemented in PyTorch~\citep{paszke2019pytorch} on an NVIDIA A100 GPU. Models were trained with Adam~\citep{kingma2014adam}, an initial learning rate of $10^{-4}$, batch size 4, and no learning-rate scheduler. Each variant was trained for 8--20 epochs, and the checkpoint with the best validation Pearson correlation was selected. Table~\ref{tab:settings} summarizes the main settings.

\begin{table}[t]
\centering
\caption{Main implementation settings.}
\label{tab:settings}
\begin{tabular}{ll}
\toprule
Parameter & Value\\
\midrule
Optimizer & Adam\\
Learning rate & $1\times10^{-4}$\\
Batch size & 4\\
Clip length & 64 / 128\\
Input size & 96$\times$96 / 128$\times$128\\
Loss & Negative Pearson Loss\\
GPU & NVIDIA A100\\
Framework & PyTorch\\
\bottomrule
\end{tabular}
\end{table}

\subsection{Evaluation Metrics}
We report Pearson correlation, heart-rate MAE estimated via FFT peak analysis, subject-level winrate, paired Wilcoxon signed-rank tests between baseline and masked conditions~\citep{wilcoxon1945individual}, and runtime throughput in frames per second (FPS).

\subsection{Physiological Reconstruction Results}
Table~\ref{tab:chrom} shows the effect of static lower-face suppression in classical CHROM. Static masking improves waveform correlation by 0.0659 Pearson and reduces heart-rate MAE by 10.06 bpm. The improvement is consistent across subjects, with an 88.1\% winrate and Wilcoxon $p=2.0\times10^{-7}$.

\begin{table}[t]
\centering
\caption{CHROM performance with and without static lower-face suppression.}
\label{tab:chrom}
\begin{tabular}{lcc}
\toprule
Method & Pearson $\uparrow$ & HR MAE $\downarrow$\\
\midrule
CHROM & 0.3507 & 33.50\\
CHROM + Static Mask & 0.4166 & 23.44\\
\bottomrule
\end{tabular}
\end{table}

Table~\ref{tab:variants-results} summarizes SKIA-Net variants. The masked-only variant has slightly lower waveform correlation than the raw-only configuration but lower HR MAE, indicating improved heart-rate stability. Combining raw and masked representations improves both metrics over the raw-only baseline. The full configuration achieves the best performance by adding complementary short-term motion cues.

\begin{table}[t]
\centering
\caption{Comparison between SKIA-Net variants.}
\label{tab:variants-results}
\begin{tabular}{lcc}
\toprule
Method & Pearson $\uparrow$ & HR MAE $\downarrow$\\
\midrule
PhysNet-lite~\citep{yu2019remote} & 0.2350 & --\\
SKIA Raw & 0.6320 & 8.99\\
SKIA Masked & 0.6246 & 7.55\\
SKIA Raw+Masked & 0.6844 & 4.97\\
SKIA Full & 0.7522 & 3.73\\
\bottomrule
\end{tabular}
\end{table}

\subsection{Ablation Studies}
Table~\ref{tab:ratio} shows the effect of the suppression ratio. Aggressive suppression ($\alpha=0.5$) removes useful skin regions, while weak suppression ($\alpha=0.8$) leaves motion-contaminated regions intact. The best balance is achieved at $\alpha=0.7$.

\begin{table}[t]
\centering
\caption{Ablation study on lower-face suppression ratio.}
\label{tab:ratio}
\begin{tabular}{ccc}
\toprule
Ratio & Mean $\Delta$ Pearson & Winrate\\
\midrule
0.5 & -0.0210 & 40.5\%\\
0.6 & +0.0592 & 81.0\%\\
0.7 & +0.0659 & 88.1\%\\
0.8 & +0.0410 & 85.7\%\\
\bottomrule
\end{tabular}
\end{table}

We further compare optical-flow-based dynamic masking with static masking in Table~\ref{tab:dynamic}. Static masking improves correlation, whereas dynamic masking degrades it. We attribute this to temporally varying spatial weights, which introduce unstable frame-wise color sampling distributions and artificial fluctuations unrelated to cardiovascular activity.

\begin{table}[t]
\centering
\caption{Static and dynamic masking strategies.}
\label{tab:dynamic}
\begin{tabular}{lccc}
\toprule
Method & Pearson $\uparrow$ & $\Delta$ vs CHROM & Winrate\\
\midrule
CHROM Baseline & 0.3507 & -- & --\\
Static Mask & 0.4166 & +0.0659 & 88.1\%\\
Dynamic Soft Mask & 0.1534 & -0.1973 & 4.8\%\\
\bottomrule
\end{tabular}
\end{table}

\subsection{Further Analysis}
\begin{figure}[t]
    \centering
    \includegraphics[width=0.92\linewidth]{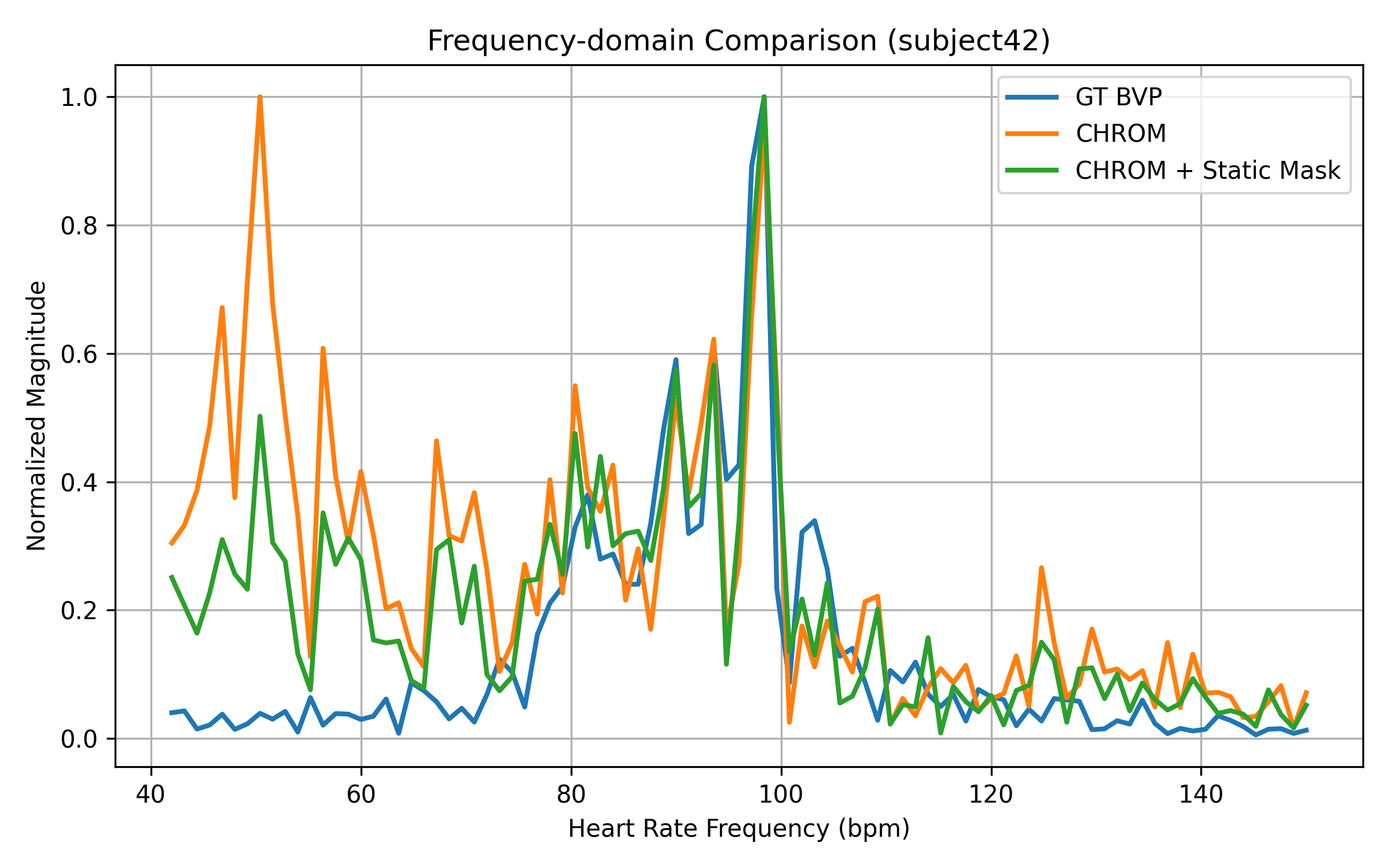}
    \caption{Frequency-domain comparison between baseline CHROM and static lower-face suppression for Subject 42.}
    \label{fig:spectrum}
\end{figure}

Figure~\ref{fig:spectrum} shows a representative frequency-domain comparison for Subject 42. Static masking suppresses spurious low-frequency peaks around 45--65 bpm while strengthening the dominant peak near the ground-truth cardiac frequency.

Runtime throughput is reported in Table~\ref{tab:runtime}. Deep-model throughput was measured on preloaded clips using GPU inference only, excluding preprocessing and data loading.

\begin{table}[t]
\centering
\caption{Runtime throughput comparison.}
\label{tab:runtime}
\begin{tabular}{lc}
\toprule
Method & FPS\\
\midrule
CHROM & 568.73\\
CHROM + Static Mask & 305.63\\
SKIA Raw & 4868.11\\
SKIA Raw+Masked & 17523.12\\
SKIA Full & 10626.73\\
\bottomrule
\end{tabular}
\end{table}

\begin{figure}[t]
    \centering
    \includegraphics[width=\linewidth]{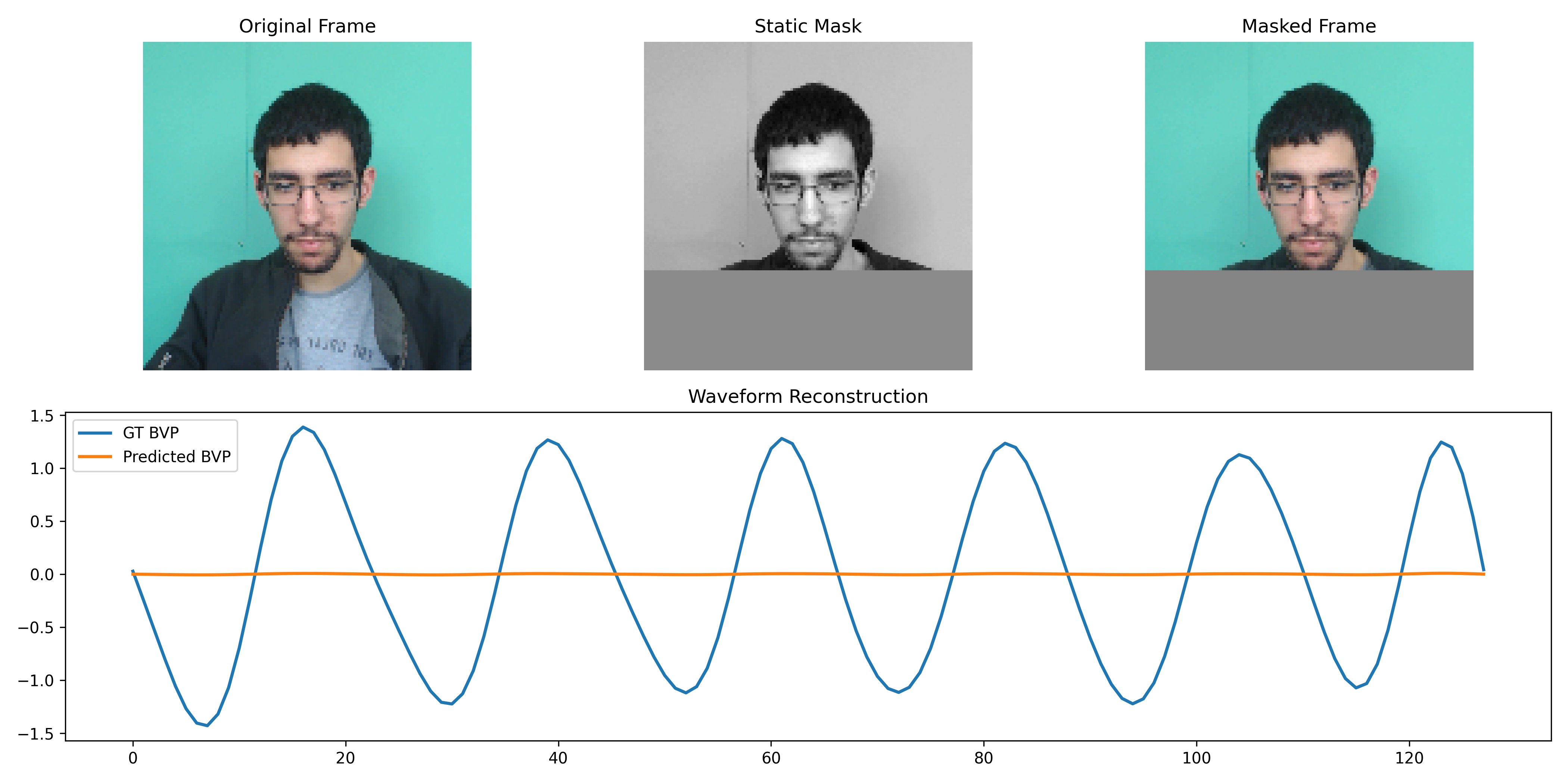}
    \caption{Qualitative visualization of SKIA-Net reconstruction. Top: original frame, static lower-face mask, and masked frame. Bottom: reconstructed BVP waveform compared with ground-truth physiological signal.}
    \label{fig:qualitative}
\end{figure}

Figure~\ref{fig:qualitative} presents a representative reconstruction example. The predicted waveform closely follows the temporal structure of the ground-truth BVP signal while remaining robust to lower-face motion.

\section{Discussion}
\subsection{Interpretation of Results}
The results consistently indicate that lower-face regions introduce structured, non-physiological variation that degrades rPPG. For CHROM, static lower-face suppression raises Pearson correlation from 0.3507 to 0.4166 and reduces HR MAE by more than 10 bpm, with an 88.1\% subject-level winrate and strong statistical significance. Deep models that incorporate masked representations also outperform single-stream baselines.

A central finding is that a simple static anatomical prior can substantially enhance reconstruction quality. The mask remains temporally stable, preserves physiologically rich upper-face regions, and reduces the influence of speaking-related intensity changes on aggregated RGB traces. In contrast, optical-flow-based dynamic masking degrades performance because frame-wise spatial weights introduce unstable color sampling distributions.

The full SKIA-Net configuration jointly combines raw appearance, masked representations, and temporal differences. The raw stream retains appearance cues, the masked stream injects an anatomical prior, and the temporal-difference stream provides complementary short-term motion cues. Their fusion enables selective integration of physiological information while reducing sensitivity to motion artifacts.

\subsection{Limitations and Future Work}
All experiments were conducted exclusively on UBFC-rPPG, so cross-dataset generalization remains to be verified. The static mask is heuristic and does not adapt to individual facial geometry; landmark-aware or segmentation-aware approaches may improve it. UBFC-rPPG is not a dedicated speaking dataset, so the speaking-related interpretation should be viewed as a motion-oriented motivation rather than a validated speech-analysis claim. Reported deep-model runtimes exclude preprocessing and data-loading overhead, and the dynamic masking study considered only a basic optical-flow implementation.

Future work will pursue cross-dataset evaluation and adaptive, face-landmark-guided region suppression. We also plan to extend the framework to speaking-intensive applications such as oral assessment, interviews, conversational interaction, and cognitive stress analysis.

\section{Conclusion}
This paper demonstrated that lower-face suppression is a simple yet effective spatial prior for robust rPPG under facial motion artifacts. On UBFC-rPPG, static lower-face masking improved CHROM correlation from 0.3507 to 0.4166 and reduced HR MAE from 33.50 to 23.44 bpm, with an 88.1\% subject-level winrate and strong statistical significance. SKIA-Net combines raw appearance, masked representations, and temporal differences; its full configuration achieved Pearson correlation 0.7522 and HR MAE 3.73 bpm. Ablation studies showed that stable anatomical suppression substantially outperforms naive optical-flow-based dynamic masking. The proposed framework improves classical and learned rPPG pipelines while remaining lightweight and interpretable.

\clearpage
\bibliographystyle{plainnat}
\bibliography{references}

\end{document}